# An Usage Measure Based on Psychophysical Relations

## Victor Kromer

It is accepted to rank words on absolute (or proportional) frequency, when making frequency lists of words, occuring in the texts. At the same time, the researchers, feeling imperfection of frequency as unconditional indicator of word "importance", consider also other characteristics of usage for revealing the basic zone of frequency list, for example the range, i.e. the number of categories, in which the word occured at least once.

However linear character of the basic dictionary assumes word ranking on only one parameter, and that has resulted in creation of so-called "distributive" dictionaries, where words are ranked only on the range (Andreev, 1967). Both these approaches can be generalized with introduction of a certain usage measure $M$, expressed through the total corpus frequency $F$ and the range $t$.

$$M = F^{1-a} t^a .  \qquad (1)$$

In the above formula $a$ stands for a certain parameter, taking values between 0 and 1 and describing the dictionary's maker distancing from the "classical" technique, with which only frequency is taken into account. It can readily be seen, that for frequency dictionary $a = 0$, and for distributive one $a = 1$. The middle value $a = 0.5$ results in a known technique of selection words on the product of frequency by the range, as $F^{1-0.5} t^{0.5} = (Ft)^{0.5}$, since raising $Ft$ to a positive power does not change the order of ranking. Such dictionaries we propose to name equidistant ones. The calculation of so-called statistical stability factor of the term (word), equal to the product of the proportional frequency of the word in the corpus by relative entry of the word (Marusenko, 1983, p. 87), leads to the latter case, where the ratio of categories, in which the word was observed, to total number of categories, is accepted for relative entry of the word.

A variant of word ranking on product of frequency by the square root from the range leads to expression (1) with $a = 1/3$, since

$$M = F^{2/3} t^{1/3} = \left(F\sqrt{t}\right)^{2/3} . \qquad (2)$$

In any case the choice of parameter $a$ value in the formula (1) reflects author's taste and his measure of choosing between two extreme variants – frequency dictionary and distributive one.

It was offered also to apply as a word usage measure the product of total corpus frequency by "coefficient of dispersion". That technique was used by A. Juillard for making up a series of frequency dictionaries of Romance languages. A number of Juillard's usage coefficient disadvantages is revealed by J. Carroll (1970/72, p. 61 – 62). Let's specify 3 of them:

1. The usage coefficient depression of a word is disproportionate the degree of its "horizontal" distribution deviation from the uniform one.

2. The Juilland's technique assumes equal category sizes, that is not always acceptable by virtue of circumstances.

3. With concentration of all frequencies in one single category Juilland's usage coefficient $U$ takes the value 0 without any dependence from actual word frequency, and that seems unreasonable.

Carroll (1970/72) offered the technique of calculating so-called "an alternative to Juilland's usage coefficient for lexical frequencies" $U_m$, based on replacement of absolute frequencies for proportional ones, and that has allowed to operate with unequally-sized categories, and replaced the coefficient of dispersion $D$ by index $D_2$, based on the information measure and equal

$$D_2 = \frac{-\Sigma p_j \log p_j}{\log n}, \qquad (3)$$

where $p_j$ is proportional to the probability of an event in the $j$-th category and is normalized proceeding from the condition $\Sigma p_j = 1$, and $n$ = number of categories.

To exclude an opportunity of coefficient $U_m$ acceptance of value 0 with concentration of all frequencies in a single category, the value of $U_m$ is scaled from a certain greater-than-zero "minimum value", proceeding from horizontal distribution of category frequencies, up to the maximum value.

On our sight, all considered above usage measures have one common disadvantage: their introduction and application is not based psycholinguistically. A usage measure, free from the specified disadvantage, is offered below.

Kondratjeva (1964, p. 16; 1972, p. 40) expressed an assumption, that psychophisical Weber-Fechner's law, valid for elementary kinds of feelings (sight, hearing, touch) can be distributed to more complex kinds of mental activity, for example to perception and memorizing of unfamiliar lexicon during reading of foreign texts.

We shall understand set of all presentations of particular words during reading the completed text as a set of external signals. In this case the potential range of perceived signals (area of adequate reflection) is limited in a natural way: the lower bound – by presentation of *hapax legomena*, i.e. words occuring in the text once, and the upper bound – by the most frequent word-type. The step of the stimul row is also determined in a natural way by presentation of a word.

Basic psychophysical Weber-Fechner's law establishes correspondence between an external signal-stimulus $S$ and answer-back reaction $R$ (Zabrodin and Lebedev, 1977, p. 75):

$$R = a \ln S + b. \tag{4}$$

The given law is valid only in a certain determined interval of external signals, and refuses in the field of small and large signals. In the field of small signals the following relations are natural for the modality, we are interested in, (consecutive presentation of the same word-type during deployment of the text): $R(0) = 0$; $R(1) = 1$. To coordinate these relations to expression (4), valid in the field of middle signals by definition, it is enough to put $a = 1$; $b = 0.5772 \ldots$ (Euler's constant), and the dependence of reaction in function of the stimulus (word with frequency $F$) will be defined by the following expression:

$$R = \sum_{k=1}^{F} \frac{1}{k}, \tag{5}$$

where $F$ = word frequency in the text. The expression (5) asymptotically approaches ($\ln F + C$) with large $F$, where $C = 0.5772 \ldots$ – Euler's constant. The hypothesis, according to which perception of a word (the reaction) is proportional to special function of word frequency (SFWF) with accumulation of separate words SFWF in the process of text deployment, is put forward by Kromer (1997). Under SFWF we understood the sum of the first $F$ members of the harmonic row $1 + \frac{1}{2} + \frac{1}{3} + \ldots + \frac{1}{F}$, where $F$ stands for the word frequency. The expression (5) can be determined through the psi-function (the logarithmic derivative of an Euler's integral of the second kind):

$$R = \psi(F+1) + C, \tag{6}$$

This expression is as well valid with $F = 0$, as $\psi(1) = -C$. The expression (6) in the field of small $F$ approximately corresponds to other known psychophisical law – Steven's power law (Zabrodin and Lebedev, 1977, p. 81):

$$R = aS^n + b, \qquad (7)$$

with parameters $a = 1$; $b = 0$; $n = 0.5$. The relations between the Steven's law, offered here dependence $R(F)$ and Weber-Fechner's law are depicted graphically in the Fig. 1. The dependences are calculated proceeding from stated above parameter values. According to (Zabrodin and Lebedev, 1977, p. 270), both psychophysical laws may not contradict each other by admittion that they are valid for different sites of feelings scale, as it may be seen from the plots in Fig. 1.

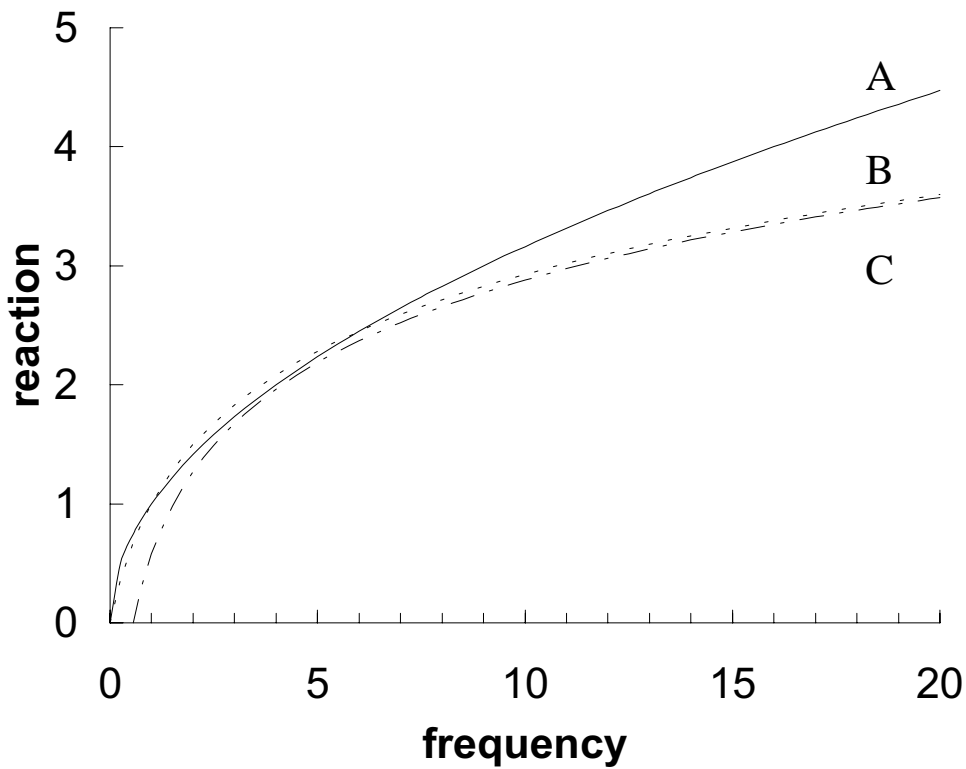

A: plot of Steven's dependence $R(S)$;
B: plot of proposed here dependence $R(F)$;
C: plot of Weber-Fechner's dependence $R(S)$.

Fig. 1.

It is accepted as a working hypothesis, that while reading the text, the subjective feeling, caused by a specific word with frequency $F$, is determined by the formula (6). It is rational in such a case to include in the basic dictionary words with the maximal sum of *R*-values in all texts of the corpus, i.e. it is offered to accept expression

$$U_R = \sum_{j=1}^{n} \left( \psi(F_j + 1) + C \right) \quad (8)$$

as a word-usage measure, where *n* = the number of the texts in the corpus, $F_j$ = word-frequency in text *j*.

Advantage of the offered measure is the simplicity of pooling several dictionaries or enlarging the text corpus, for what there is enough to sum the $U_R$-values for each word in dictionaries under considerations, or to add $U_R$, specified for added texts and arrange the new list in accordance with decreasing $U_R$.

Let's compare $U_R$-values with Juilland's and Carroll's usage coefficients, for what we put in table 1 the illustrative hypothetical data given by Juilland and Carroll (Carroll, 1970/72, p. 63) and compare efficiency of the offered measure with Juilland's and Carroll's usage coefficients.

**Table**

| Word | Frequencies by category | | | | | Total | U | $U_m$ | $U_R$ |
|------|---|---|---|---|---|-------|------|-------|-------|
|      | A | B | C | D | E |       |      |       |       |
| 1 | 1 | 1 | 1 | 1 | 1 | 5  | 5,00 | 5,00 | 5,00 |
| 2 | 2 | 1 | 1 | 1 | 0 | 5  | 3,42 | 4,31 | 4,50 |
| 3 | 2 | 2 | 1 | 0 | 0 | 5  | 2,76 | 3,62 | 4,00 |
| 4 | 3 | 1 | 1 | 0 | 0 | 5  | 2,26 | 3,36 | 3,83 |
| 5 | 3 | 2 | 0 | 0 | 0 | 5  | 1,84 | 2,67 | 3,33 |
| 6 | 4 | 1 | 0 | 0 | 0 | 5  | 1,13 | 2,24 | 3,08 |
| 7 | 5 | 0 | 0 | 0 | 0 | 5  | 0,00 | 1,00 | 2,28 |
| 8 | 0 | 0 | 3 | 3 | 4 | 10 | 5,82 | 7,41 | 5,75 |
| 9 | 1 | 1 | 1 | 1 | 6 | 10 | 5,00 | 8,10 | 6,45 |

Words 1 through 7 exibit the seven possible distributions of five frequencies over five categories (assuming category identities are irrelevant). It will be observed that the values of $U$, $U_m$ and $U_R$ decrease from the maximal value down to the minimal one as all frequencies concentrate in a single category.

On words 8 and 9 Juilland demonstrates the incorrectness of the offered usage coefficient on some occassions. So, on comparing word 9 and word 1, common sense suggests, that word 9 is more "important", since it with equal frequencies in categories A – D has larger frequency in category E, however $U$-value is in both cases the same. Carroll's usage coefficient $U_m$ eliminates this injustice and simultaneously changes the order of words 8 and 9 in rank distribution, and that corresponds to the importance of these words according to the submitted data, however values of $U_m$ for these words seem to be overestimated.

The offered in the present paper usage measure $U_R$ , owing to its psycholinguistic substantiation, is, in author's opinion, very acceptable for estimating the degree of "word-importance" and making the dictionaries of basic languages on the quantitative characteristics.

Russian writer of the 19-th century N. Gogol's work in 2 volumes "Evenings on a Farm near Dikanka" (8 narratives and 2 forewords) was statistically processed for testing the offered usage measure efficiency. The language of the texts is Russian. Word tokens served as registration units, hyphen was considered as a letter. The total corpus was 62,358 word-tokens. It is accepted to be guided by a reliable frequency threshold equal 35 for an estimation of the reliable zone of the frequency dictionary (Alekseev, 1975, p. 52). The frequency dictionary of the whole work contains 197 words with frequency equal or greater than 35.

In the reliable zone of the dictionary there were occuring many proper names and nominal words carrying out function of proper names. At the same time the reliability of frequencies is based on the certain model and is of a relative character for real texts (Alekseev, 1975, p. 45). The basic assumption is that all parts of the text are arranged under one and the same law. The said hypothesis is denied by taking into account even the most frequent word-types, the probabilities of which are estimated on frequencies more precisely than those of other words (Arapov, 1988, p. 17), and the phenomenon of "huddling together" of rare words (Boroda et al., 1977). All this taken together allows to consider the estimation of the reliable zone of the frequency dictionary on threshold frequency 35 idealized and not valid for real texts. At the same time refusal from total frequency as the basic criterion of word selection and taking into account horizontal distribution peculiarities allows to come nearer to the considered ideal case.

It is offered in the present paper to use the sum of *R*-values, (i.e. the proposed usage measure $U_R$), calculated on the formula (6) on all categories for word selection for lexical minima, basic languages etc. (categories should be completed texts).

The results of processing by the offered technique the work "Evenings on a farm near Dicanka" are given below. From the word-type list, ranked on $U_R$, also 197 top ranking words were selected.

Comparison of two lists (with word selection on frequency and on $U_R$) reveals a common part (169 words). The analysis of frequency list allows to make a conclusion, that a significant part from 28 words which were not included in the list of words, ranked on $U_R$, are proper names and nominal words in function of proper names. These words have come into the top zone of the frequency dictionary quite accidentally and will be taken away when enlarging the text corpus, because "when the sample is multiplied and more and more texts are included into it, then one can easily see the common things typical of the hypothetical majority of such texts. In case the samples are very big, the words without which no large text can do, come to the top zone of the frequensy dictionary. First of all they are functional words and those notional ones which are most widely spread, stylistically neutral, most compatible, polysemantic, etc., i.e. those which are considered to be learnt first by traditional linguadidactics" (Alekseev, 1988, p. 136).

However in some cases the opportunity of the text corpus expansion is either limited, or excluded owing to the corpus closeness. The word selection basing on usage measure $U_R$ allows to reveal words, subjected to selection and mastering by students, from a corpus of rather modest size, as the irregularities in horizontal distribution, caused by casual reasons of extralinguistic character will be suppressed.

It seems to us, that the acceptance of the hypothesis about dependence between perception (reaction) and stimulus by the formula (6) with accumulation of separate $R(F)$ values of separate completed texts allows to come nearer to understanding of idiolect (speech and mental activity of a person) formation mechanism. With all randomness and limitations of person's "text corpus" (understood here in a very wide sense) by the time of his idiolect formation, the considered mechanism reveales from "the corpus" common and stylistically neutral words, which are reproduced then by the person under consideration most frequently and quite evenly, so giving contribution to formation of other persons "text corpus".

Thus, national language appears enveloped by a system of positive feedback loops, and that can be considered from the position of linguistic sinergetics as internal self-organizing and self-regulating mechanisms manifestation. As it is known, a system, enveloped with a positive feedback loop, is inclined to self-excitation under certain conditions, and the presence of several feedback contours may result in "trigger" effects, i.e. sudden change of system parameters caused by external factor.

It is obvious, that feedback parameters and the $R(F)$-dependence parameters take in process of "tuning" optimal values from national language functioning point of view. So, the smaller degree of positive feedback results in eroding of borders between common and special vocabulary, what would complicate communication between different idiolect carriers, and the absence of dependence $R(F)$ "compression" with the given feedback parameters would lead to system self-excitation and "trigger" effects.

It is possible also, that the separate person speech-mental activity disorder may be explained by deviating feedback loops parameters values and dependences "stimulus – reaction".

The considered in the present paper technique of word selection on usage measure $U_R$ is supposed to be used for making educational dictionaries-minima, basic dictionaries of sublanguages, and also for selection of lexicon, subjected for active mastering during educational reading.